%% file: iclr2026_conference.tex
\documentclass{article} 
\usepackage{iclr2026_conference,times}

\input{math_commands.tex}

\usepackage{graphicx}
\usepackage{hyperref}
\usepackage{url}
\usepackage{tikz}
\usetikzlibrary{arrows.meta, positioning}

\title{Is Evaluation Awareness Just Format Sensitivity? Limitations of Probe-Based Evidence under Controlled Prompt Structure}

\author{Viliana Devbunova \thanks{alternative address: devbunova99@gmail.com} \\
Yandex \\
Belgrade, Serbia \\
\texttt{viliana-dev@yandex-team.ru} \\
}

\iclrfinalcopy 
\begin{document}

\maketitle

\begin{abstract}
Prior work uses linear probes on benchmark prompts as evidence of \emph{evaluation awareness} in large language models. Because evaluation context is typically entangled with benchmark format and genre, it is unclear whether probe-based signals reflect context or surface structure. We test whether these signals persist under partial control of prompt format using a controlled $2\times2$ dataset and diagnostic rewrites. We find that probes primarily track benchmark-canonical structure and fail to generalize to free-form prompts independent of linguistic style. Thus, standard probe-based methodologies do not reliably disentangle evaluation context from structural artifacts, limiting the evidential strength of existing results.
\end{abstract}

\section{Introduction}

Recent work introduces the notion of \emph{evaluation awareness} in large language models (LLMs): the ability to distinguish evaluation-time prompts from deployment-like usage and potentially modulate behavior accordingly \citep{needham2025largelanguagemodelsknow, chaudhary2025evaluationawarenessscalespredictably}. This phenomenon has implications for benchmark validity and safety monitoring, as evaluation-time behavior may not reflect real-world deployment \citep{hua2026steeringevaluationawarelanguagemodels}.

\begin{figure}[b]
\centering
\begin{tikzpicture}[
    box/.style={draw, rounded corners=2pt, minimum height=0.5cm, inner sep=3pt, font=\scriptsize},
    lbl/.style={font=\scriptsize\itshape, text=black!60},
    arrow/.style={-{Stealth[length=1.5mm]}, semithick},
    every node/.style={font=\scriptsize}
]

\node[font=\scriptsize\bfseries, anchor=west] at (-0.2, 2.0) {(a) Standard probe setup};

\node[box, fill=blue!10, minimum width=1.4cm] (bench) at (0.7, 1.15) {Benchmark};
\node[lbl, anchor=west] at (1.9, 1.15) {structured + eval};

\node[box, fill=orange!10, minimum width=1.4cm] (chat) at (0.7, 0.2) {Chat};
\node[lbl, anchor=west] at (1.9, 0.2) {free-form + deploy};

\node[box, fill=gray!8, minimum width=0.9cm] (llm) at (4.6, 0.675) {LLM};
\node[box, fill=gray!8, minimum width=0.9cm] (probe) at (5.9, 0.675) {Probe};

\node[font=\scriptsize, text=blue!60] (eout) at (7.1, 1.05) {Eval};
\node[font=\scriptsize, text=orange!60] (dout) at (7.1, 0.3) {Deploy};

\draw[arrow] (bench.east) -- ++(0.55,0) |- (llm.west);
\draw[arrow] (chat.east) -- ++(0.55,0) |- (llm.west);
\draw[arrow] (llm) -- (probe);
\draw[arrow] (probe.east) -- ++(0.1,0) |- (eout.west);
\draw[arrow] (probe.east) -- ++(0.1,0) |- (dout.west);

\draw[dashed, red!50, semithick, rounded corners=3pt]
    (-0.15, -0.15) rectangle (3.8, 1.55);
\node[font=\tiny\itshape, text=red!60, anchor=north] at (1.825, -0.2) {format $\approx$ context (confound)};

\node[font=\scriptsize\bfseries, anchor=west] at (8.2, 2.0) {(b) Our controlled design};

\def\ox{9.2}
\def\oy{0.675}
\def\cw{1.55}
\def\ch{0.55}

\node[font=\tiny\bfseries] at (\ox+0.5*\cw, \oy+\ch+0.22) {Eval};
\node[font=\tiny\bfseries] at (\ox+1.5*\cw, \oy+\ch+0.22) {Deploy};
\node[font=\tiny\bfseries, rotate=90, anchor=south] at (\ox-0.4, \oy+0.6*\ch) {Bench};
\node[font=\tiny\bfseries, rotate=90, anchor=south] at (\ox-0.4, \oy-0.6*\ch) {Casual};

\fill[blue!8]   (\ox,      \oy)      rectangle (\ox+\cw,   \oy+\ch);
\fill[red!8]    (\ox+\cw,  \oy)      rectangle (\ox+2*\cw, \oy+\ch);
\fill[green!8]  (\ox,      \oy-\ch)  rectangle (\ox+\cw,   \oy);
\fill[orange!8] (\ox+\cw,  \oy-\ch)  rectangle (\ox+2*\cw, \oy);

\draw[semithick] (\ox, \oy-\ch) rectangle (\ox+2*\cw, \oy+\ch);
\draw[semithick] (\ox+\cw, \oy-\ch) -- (\ox+\cw, \oy+\ch);
\draw[semithick] (\ox, \oy) -- (\ox+2*\cw, \oy);

\node[font=\tiny] at (\ox+0.5*\cw, \oy+0.5*\ch) {Bench-Eval};
\node[font=\tiny] at (\ox+1.5*\cw, \oy+0.5*\ch) {Bench-Deploy};
\node[font=\tiny] at (\ox+0.5*\cw, \oy-0.5*\ch) {Casual-Eval};
\node[font=\tiny] at (\ox+1.5*\cw, \oy-0.5*\ch) {Casual-Deploy};

\node[font=\tiny\itshape, text=black!60, anchor=north] at (\ox+\cw, \oy-\ch-0.15) {format $\perp$ context};
\end{tikzpicture}
\caption{Overview of our approach. \textbf{(a)}~Standard probes train on benchmark vs.\ chat prompts, where format and context are confounded. \textbf{(b)}~Our $2\times2$ design crosses format and context independently, enabling isolation of each factor.}
\label{fig:overview}
\end{figure}
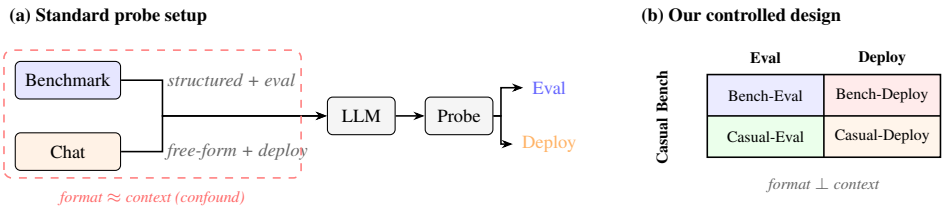

Empirical evidence for evaluation awareness largely relies on \emph{probe-based analyses}, where linear classifiers trained on frozen activations distinguish evaluation from deployment prompts \citep{nguyen2025probingsteeringevaluationawareness}. However, probe accuracy is not diagnostic: classifiers may exploit any feature correlated with labels. In practice, evaluation prompts are tightly coupled with benchmark-canonical formats and neutral genres, while deployment prompts are more conversational \citep{needham2025largelanguagemodelsknow}. This coupling raises the concern that probes track surface structure rather than evaluation context itself (Figure~\ref{fig:overview}a).

We ask a methodological question: \emph{Do probe-based signals attributed to evaluation awareness remain informative once prompt format and genre are partially controlled?} Rather than testing whether evaluation awareness exists, we assess whether standard probe-based methods can isolate evaluation context under stricter controls.

Prior work reports cross-dataset generalization of evaluation-awareness probes, but existing datasets rarely break the correlation between evaluation context and benchmark-canonical format. We introduce controlled format-transplant and genre controls showing that, under the standard single-contrast training setup, probe signals largely track benchmark structure rather than context.

We make three contributions:
\begin{itemize}
\item \textbf{Controlled 2$\times$2 design.} We construct datasets and diagnostic rewrites that independently vary context, genre, and format under matched prompt length (Figure~\ref{fig:overview}b).
\item \textbf{Probe generalization fails under format-transplant interventions.} Linear probes trained on benchmark contrasts fail to generalize to free-form prompts, even when content and style are preserved.
\item \textbf{Methodological implications.} Probe-based evaluation-awareness diagnostics are highly format-sensitive, limiting their use as structure-invariant safety tools.
\end{itemize}

\section{Related Work}

Several recent studies examine whether LLMs internally distinguish evaluation from deployment settings. \citet{needham2025largelanguagemodelsknow} show that state-of-the-art models can classify transcripts as evaluation or deployment with high accuracy (AUC $\approx$ 0.83, compared to a human upper bound of $\approx$ 0.92), suggesting that prompts contain systematic cues associated with evaluation contexts.

\citet{nguyen2025probingsteeringevaluationawareness} probe frozen internal activations using linear classifiers and report generalization across held-out datasets. They observe that safety-related benchmarks are consistently flagged as evaluation prompts even without explicit training, indicating that benchmark structure itself may act as a strong cue. Relatedly, \citet{hua2026steeringevaluationawarelanguagemodels} train a synthetic “organism” model to exhibit evaluation-aware behavior and identify an activation direction whose removal suppresses such behavior, demonstrating that evaluation-linked representations can be induced and manipulated.

\citet{chaudhary2025evaluationawarenessscalespredictably} study how probe-based evaluation awareness scales with model size, reporting a power-law relationship across open-weight models ranging from 270M to 70B parameters. Together, these results suggest that probe-detectable signals associated with evaluation settings are widespread and scale systematically.

Our work differs by testing whether such probe-based signals persist under stricter controls on prompt format and genre. Rather than asking whether evaluation awareness exists, we assess whether standard probe-based diagnostics isolate evaluation context or instead reflect sensitivity to benchmark-canonical structure.

\section{Setup and Methods}
\label{sec:methods}

We investigate whether probe-based evaluation awareness is a genuine response to usage context or an artifact of sensitivity to prompt format. Unlike prior work that conflates these dimensions, our design treats \emph{Context} (Evaluation vs. Deployment) and \emph{Format} (Benchmark vs. Casual) as independent variables.

\subsection{The 2x2 Controlled Design}
We construct a dataset matrix covering four quadrants to isolate the effects of structure. We define our variables as follows:
\begin{itemize}
    \item \textbf{Context:} The underlying intent. \emph{Evaluation} tests capabilities; \emph{Deployment} assists a user.
    \item \textbf{Format:} The surface structure. \emph{Benchmark} implies rigid templates (e.g., MCQ); \emph{Casual} implies free-form inputs.
\end{itemize}

\begin{table}[h]
\caption{The $2\times2$ Dataset Matrix. We cross usage context with prompt format to disentangle structural cues from evaluation intent.}
\label{tab:datasets}
\begin{center}
\begin{tabular}{l|c|c}
\multicolumn{1}{c|}{\bf Format} & \multicolumn{1}{c|}{\bf Evaluation Context} & \multicolumn{1}{c}{\bf Deployment Context} \\
\hline
\textbf{Benchmark} & \textbf{Bench-Eval} & \textbf{Bench-Deploy} \\
(Structured) & (MMLU, ARC, GSM8K) & (ShareGPT rewritten to \\
 & \textit{Standard Evaluation} & benchmark format) \\
\hline
\textbf{Casual} & \textbf{Casual-Eval} & \textbf{Casual-Deploy} \\
(Free-form) & (MuTual, 1st turn) & (ShareGPT, 1st turn) \\
 & \textit{Conversational Proxy} & \textit{Standard Deployment} \\
\end{tabular}
\end{center}
\end{table}

\paragraph{Dataset Definitions.}
\textbf{Bench-Eval} consists of canonical prompts from MMLU, ARC, and GSM8K. \textbf{Casual-Deploy} uses deployment-style user prompts sampled from ShareGPT (restricted to the first turn). \textbf{Casual-Eval} acts as a conversational evaluation proxy, using the first turn of the MuTual dataset. Finally, \textbf{Bench-Deploy} serves as an adversarial control: we route ShareGPT queries to compatible tasks and rewrite them into strict benchmark formats using DeepSeek-V3 (see Appendix~\ref{app:rewrite-prompts} for prompts).

\subsection{Experimental Setup}
To ensure a clean experiment, we apply strict controls:
\begin{itemize}
    \item \textbf{Length Control:} To prevent probes from relying on length heuristics, we applied \emph{histogram matching} to align the sample length distribution of \textit{Casual-Deploy} with \textit{Bench-Eval}. Figure~\ref{fig:length_boxplot} in Appendix~\ref{app:length} illustrates the resulting distributions.
    \item \textbf{Leakage Removal:} We explicitly audited rewrites to remove artifacts such as "Answer:" fields.
    \item \textbf{Model and Probes:} We extract frozen residual stream activations (last token) from \texttt{Llama-3.1-8B-Instruct} across middle-to-late layers (indices 8--31, 0-indexed). For each layer, we train a Logistic Regression probe using 5-fold cross-validation. Detailed hyperparameters are provided in Appendix~\ref{app:implementation}.
\end{itemize}

\subsection{Training Configurations}
We compare two training paradigms with distinct evaluation protocols:
\begin{itemize}
    \item \textbf{Standard Training (Single-Contrast):} Trained on the typical confound (\textit{Bench-Eval} vs. \textit{Casual-Deploy}). We do not use a validation split here, as models trivially achieve near 100\% accuracy via format shortcuts. Instead, we evaluate \emph{zero-shot} on held-out, format-shifted datasets (OOD evaluation) to diagnose generalization failure.
    \item \textbf{Paired Training (Decorrelated):} Trained on pooled data (\textit{Bench-Eval} $\cup$ \textit{Casual-Eval} vs. \textit{Bench-Deploy} $\cup$ \textit{Casual-Deploy}). Here, we use a standard \textbf{80/20 train/test split} to demonstrate that the model can learn a generalizable context signal (high test accuracy) when format is decorrelated from the label.
\end{itemize}
\section{Observed Outcome}
\label{sec:results}

Our experiments demonstrate that standard probe-based evidence for evaluation awareness is fragile. We present results for Layer 24 (consistent trends across all layers are reported in Appendix~\ref{tab:single_full}).

\paragraph{The Format Trap.}
As shown in Table~\ref{tab:combined_results} (top row), standard probes fail to generalize. Deployment prompts rewritten into benchmark formats (\textit{Bench-Deploy}) are misclassified as ``Evaluation'' ($94.5\%$). The probe detects the structural format (e.g., MCQ), not the context. 
To isolate \textbf{Genre} (linguistic register), we tested deployment prompts rewritten into a formal ``Academic Tone'' without structural changes. This had negligible effect ($\approx 0\%$ classified as Eval), confirming that probes are sensitive to rigid templates rather than linguistic style.

\paragraph{Paired Training Success.}
When format is decorrelated from context during training, the probe successfully learns the underlying concept. As shown in Table~\ref{tab:combined_results} (bottom row), the paired probe correctly identifies \textit{Bench-Deploy} as deployment ($7.2\%$ error vs $94.5\%$) and recognizes \textit{Casual-Eval} as evaluation ($100\%$). Furthermore, we evaluated these probes on AI Safety datasets (AdvBench, HarmBench) and found that neither standard nor paired probes flag safety prompts as "Evaluation" (see Appendix~\ref{app:safety}).

\begin{table}[t]
\caption{\textbf{Probing Results (Layer 24).} Values indicate the percentage of prompts classified as \emph{Evaluation}. Standard training conflates format with context, leading to high false positives on \textit{Bench-Deploy}. Paired training restores correct classification.}
\label{tab:combined_results}
\begin{center}
\begin{tabular}{l|cc|cc}
\multicolumn{1}{c|}{} & \multicolumn{2}{c|}{\textbf{Intended Evaluation}} & \multicolumn{2}{c}{\textbf{Intended Deployment}} \\
\multicolumn{1}{c|}{\bf Training Regime} & \bf Bench-Eval & \bf Casual-Eval & \bf Casual-Deploy & \bf Bench-Deploy \\
\hline
\textbf{Standard (Single)} & \textbf{98.8\%} & 0.0\% & 0.0\% & \textbf{94.5\%} (FP) \\
\textbf{Paired (Decorrelated)} & \textbf{98.5\%} & \textbf{100.0\%} & \textbf{0.0\%} & \textbf{7.2\%} \\
\end{tabular}
\end{center}
\end{table}

\section{Reason for Failure}

\paragraph{Diagnosis.}
The observed collapse indicates that probe-based diagnostics exploit surface-level structural features rather than encoding an abstract notion of evaluation context. Benchmark-canonical layouts introduce strong, linearly separable cues in representation space, causing probes to conflate “being evaluated” with “looking like a benchmark.”

\paragraph{Why this is not a trivial confound.}
Prompt length is explicitly matched across evaluation and deployment datasets, ruling out length-based shortcuts. Genre and formality are manipulated independently from structure, yet do not induce evaluation-like predictions. We explicitly audit and remove \emph{rewrite-induced answer leakage}, where benchmark-format rewrites of deployment prompts inadvertently include model-generated answer content (e.g., pre-filled options or solution scaffolds). Ablations show that removing such leakage does not explain the observed effect and in some cases increases probe confidence. The failure is consistent across layers, indicating a pervasive structural bias rather than a localized artifact.

\paragraph{Boundary conditions.}
We do not claim that evaluation awareness is absent from model representations. Rather, our results show that under standard benchmark-correlated training setups, linear probes fail to isolate such signals. Only when format is explicitly decorrelated from context during training do probes exhibit format-invariant behavior.

\section{Discussion}

\paragraph{Implications.}
Our results suggest that probe-based monitoring of evaluation awareness can induce false confidence. Diagnostics sensitive to prompt format rather than usage context risk mischaracterizing model behavior, and governance mechanisms relying on probe accuracy may systematically overestimate situational awareness. Benchmark-canonical structure must be explicitly controlled for in experimental design, not bundled with context by default. Without such control, probe accuracy may reflect sensitivity to layout rather than higher-level contextual reasoning.

\paragraph{Limitations.}
Our study uses only linear probes and a single backbone (Llama-3.1-8B-Instruct), and does not rule out non-linear or distributed representations of evaluation awareness. Some dataset instantiations rely on proxy constructions. We expect the central confound---benchmark structure correlating with evaluation context---to persist broadly due to common benchmark construction practices.

\section*{Reproducibility Statement}

All code and constructed datasets are publicly available at \url{https://github.com/viliana-dev/eval-awareness-format}. The repository includes scripts for the full pipeline: dataset downloading, query routing, format rewrites, histogram-based length matching, activation extraction, probe training, and all evaluation protocols reported in this paper. Pre-processed datasets for the $2\times2$ matrix are hosted at \url{https://huggingface.co/datasets/viliana-dev/eval-awareness-2x2}.

All source datasets are public: MMLU \citep{hendrycks2021mmlu}, ARC \citep{clark2018arc}, GSM8K \citep{cobbe2021gsm8k}, ShareGPT\footnote{\url{https://huggingface.co/datasets/theblackcat102/sharegpt-english}}, and MuTual \citep{cui2020mutual}. Benchmark-format rewrites were generated using DeepSeek-V3; exact prompt templates are provided in Appendix~\ref{app:rewrite-prompts}.

All experiments use \texttt{meta-llama/Llama-3.1-8B-Instruct} with \texttt{bfloat16} inference on a single NVIDIA A100 80\,GB. Probes are trained in \texttt{float32}. A fixed random seed of 42 is used throughout all scripts. Implementation details---layer indices, sample sizes ($n{=}1076$), regularization, and cross-validation settings---are reported in Appendix~\ref{app:implementation}.

\bibliographystyle{iclr2026_conference}
\bibliography{iclr2026_conference}

\appendix

\section{Appendix: Datasets, Controls, and Additional Experiments}

\subsection{Datasets in the 2x2 Matrix}
\label{app:datasets}
We use a controlled dataset matrix crossing \textbf{usage context} (evaluation vs. deployment) and \textbf{genre} (benchmark-style vs. conversational), while treating \textbf{format} (benchmark-canonical structured vs. free-form) as a separate variable.

\paragraph{Bench-Eval (Evaluation; benchmark-style; structured).} Canonical benchmark prompts (e.g., MCQ and structured word-problem formats) drawn from MMLU, ARC, and GSM8K.

\paragraph{Casual-Deploy (Deployment; conversational; free-form).} Deployment-style user prompts sampled from ShareGPT. For comparability with Bench-Eval, we use only the first user turn and apply histogram matching for length control (see Section~\ref{app:length}).

\paragraph{Casual-Eval (Evaluation; conversational; free-form proxy).} We instantiate this condition with MuTual as a conversational evaluation proxy. For tighter control over genre and input structure, we extract only the first dialogue turn from each example; we treat this as an approximation rather than a faithful instantiation of the original task.

\paragraph{Bench-Deploy (Deployment; benchmark-style; structured proxy via rewrites).} Deployment-origin prompts from Casual-Deploy rewritten into benchmark-canonical structured templates (MCQ or structured word problems), while avoiding explicit evaluation framing. Since not all deployment prompts can be sensibly forced into benchmark templates, we first route prompts into coarse task categories and rewrite only compatible prompts.

\subsection{Length Control}
\label{app:length}
\paragraph{Motivation.} Initially, the datasets exhibited significant length disparities, creating a potential shortcut for probes. \textit{Bench-Eval} prompts had a median length of 248 characters, while the raw \textit{Casual-Deploy} (ShareGPT 1st turn) prompts were much shorter (median 113 characters).

\begin{figure}[t]
\begin{center}
\includegraphics[width=0.95\linewidth]{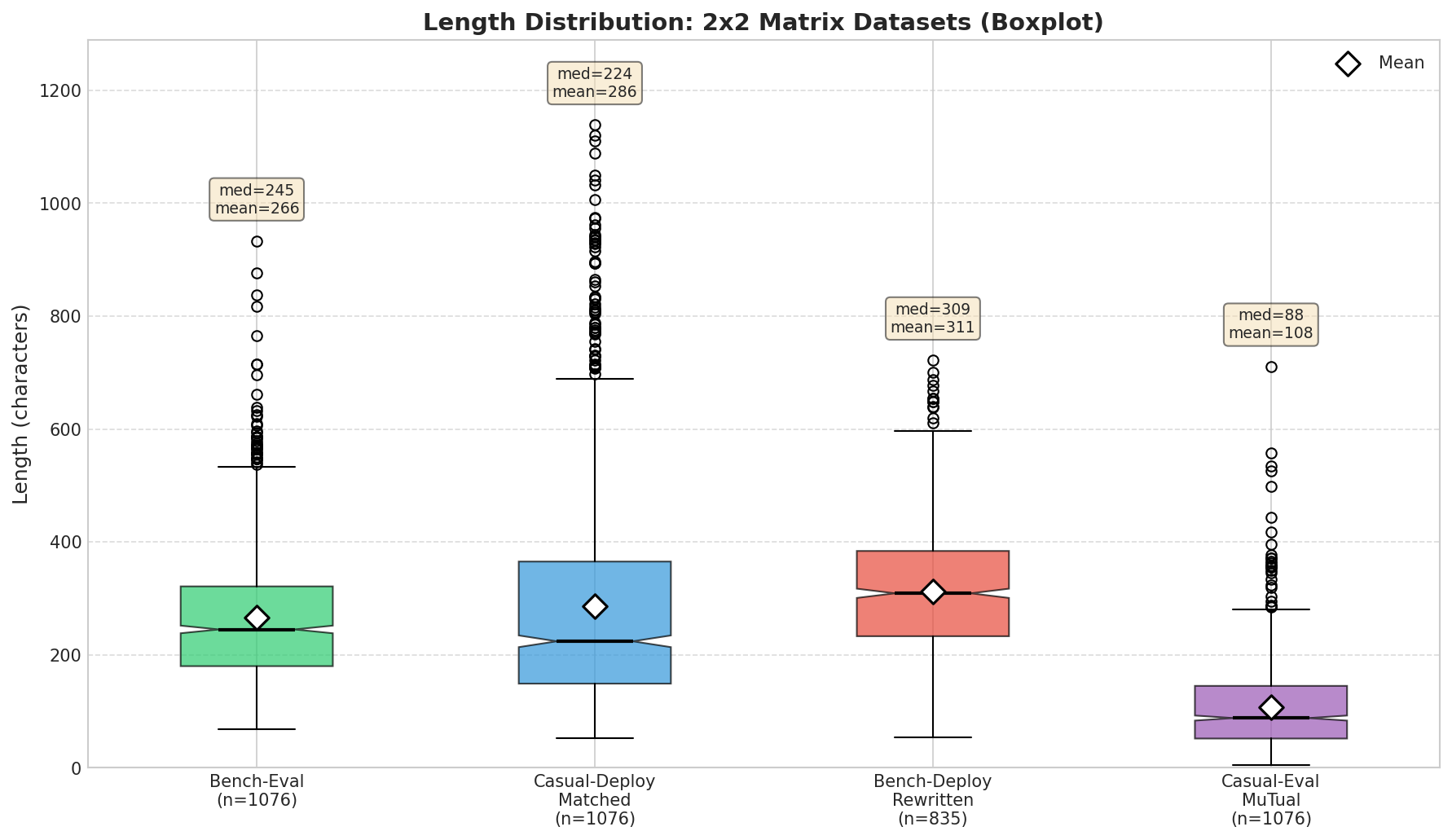}
\end{center}
\caption{
Length distributions (in characters) across the four datasets. \textit{Casual-Deploy} is histogram-matched to \textit{Bench-Eval}. \textit{Bench-Deploy} is slightly longer due to formatting overhead, while \textit{Casual-Eval} (1st turn) is naturally shorter.
}
\label{fig:length_boxplot}
\end{figure}

\paragraph{Histogram Matching.} To eliminate length as a confounding variable in the primary training contrast (\textit{Bench-Eval} vs. \textit{Casual-Deploy}), we performed histogram matching. We calculated the length distribution of \textit{Bench-Eval} using 30 percentile-based character-length bins and sampled the exact same number of examples from \textit{Casual-Deploy} for each bin. This process aligned the distributions as closely as possible, yielding equal sample sizes ($n=1076$) and comparable statistics. Figure~\ref{fig:length_boxplot} visualizes the resulting length distributions.

\subsection{Rewrite Pipeline and Leakage Removal}
\label{app:rewrites_leakage}

\paragraph{Rewrite Variants.} All rewrites are performed with DeepSeek-V3 under controlled prompting. We construct three variants sharing identical content (Casual-Deploy):
\begin{itemize}
    \item \texttt{bench\_deploy}: benchmark format + conversational genre.
    \item \texttt{casual\_deploy\_formal}: free-form format + academic/impersonal genre.
    \item \texttt{bench\_deploy\_formal}: benchmark format + academic/impersonal genre.
\end{itemize}

\paragraph{Answer Leakage Cleanup.} We define "answer leakage" as rewrite artifacts where structured templates inadvertently elicit model-generated answers (e.g., pre-filled "Answer:" fields). We explicitly audit and remove such leakage. As shown in Table~\ref{tab:leakage}, removing leakage increases the probe's confidence in classifying rewrites as "Evaluation," confirming that leakage was acting as a confounding feature that suppressed the format signal.

\begin{table}[h]
\caption{Effect of removing unintended answer leakage. Values indicate \% classified as Evaluation. Removing leakage makes the probe \emph{more} likely to flag the prompt as Eval, isolating the format effect.}
\label{tab:leakage}
\begin{center}
\begin{tabular}{lcc}
\bf Condition & \bf Before cleanup & \bf After cleanup \\
\hline
Bench-Deploy (MCQ rewrite) & 55--70\% & 94--95\% \\
Bench-Deploy Formal & 51--68\% & 88--91\% \\
\end{tabular}
\end{center}
\end{table}

\subsection{Implementation Details}
\label{app:implementation}

\paragraph{Activation Extraction.}
We extract activations using the following settings:
\begin{itemize}
    \item \textbf{Position:} Last token activation (before final layer norm) to capture the aggregated context.
    \item \textbf{Layers Evaluated:} We sweep across middle-to-late layers (0-indexed) to verify consistency: 8, 12, 16, 20, 22, 24, 26, 28, 30, 31.
    \item \textbf{Context:} Full input context is preserved (no truncation) to maintain task integrity.
    \item \textbf{Precision:} Model inference is performed in \texttt{bfloat16}; probe training uses \texttt{float32}.
\end{itemize}

\paragraph{Probe Training.}
\begin{itemize}
    \item \textbf{Architecture:} Logistic Regression (linear probe).
    \item \textbf{Preprocessing:} Input activations are normalized using \texttt{StandardScaler} fitted on the training split.
    \item \textbf{Labels:} Binary classification where 0 = Evaluation, 1 = Deployment.
    \item \textbf{Optimization:} L2 regularization (default $C=1.0$), max iterations = 1000.
    \item \textbf{Validation:} We use 5-fold Cross-Validation on the training set to verify stability.
\end{itemize}

\subsection{Additional Experimental Results}
\label{app:extended_results}

\paragraph{Standard (Single) vs. Paired Training Analysis.} 
We present the full layer-wise results for both training regimes. Table~\ref{tab:single_full} shows the failure of standard probes to generalize: they consistently misclassify \textit{Bench-Deploy} as Evaluation ($>94\%$) across all layers.
In contrast, Table~\ref{tab:paired_full} shows that probes trained on the paired dataset (decorrelating format from context) successfully learn to distinguish context across all layers.

\begin{table}[h]
\caption{\textbf{Standard (Single-Contrast) Training Results across Layers.} Values indicate \% classified as Eval. The standard probe consistently relies on format, incorrectly flagging \textit{Bench-Deploy} as Evaluation and ignoring \textit{Casual-Eval}.}
\label{tab:single_full}
\begin{center}
\begin{tabular}{c|cccc}
\bf Layer & \bf B-Eval & \bf C-Eval & \bf C-Deploy & \bf B-Deploy \\
\hline
8 & 99.8\% & 0.0\% & 0.0\% & 94.3\% \\
16 & 99.5\% & 0.0\% & 0.0\% & 94.3\% \\
24 & 98.8\% & 0.0\% & 0.0\% & 94.5\% \\
30 & 99.2\% & 0.0\% & 0.0\% & 94.6\% \\
31 & 99.1\% & 0.0\% & 0.0\% & 94.6\% \\
\end{tabular}
\end{center}
\end{table}

\begin{table}[h]
\caption{\textbf{Paired Training Results across Layers.} Values indicate \% classified as Eval. The paired probe consistently learns Context (Evaluation) vs. Format (Benchmark).}
\label{tab:paired_full}
\begin{center}
\begin{tabular}{c|cccc}
\bf Layer & \bf B-Eval & \bf C-Eval & \bf C-Deploy & \bf B-Deploy \\
\hline
8 & 98.8 & 98.6 & 0.0 & 7.2 \\
16 & 97.7 & 99.1 & 0.0 & 4.8 \\
24 & 98.5 & 100.0 & 0.0 & 7.2 \\
30 & 99.0 & 99.5 & 0.0 & 7.2 \\
31 & 98.7 & 99.5 & 0.0 & 6.0 \\
\end{tabular}
\end{center}
\end{table}

\paragraph{Evaluation on AI Safety Datasets.} 

\label{app:safety}
We evaluate probes on safety benchmarks (AdvBench, HarmBench) to test if "safety" prompts trigger evaluation awareness. As shown in Table~\ref{tab:safety_results}, safety prompts are universally classified as Deployment (0\% Eval) by both standard and paired probes. This suggests that evaluation awareness probes should not be interpreted as general-purpose safety monitors.

\begin{table}[h]
\caption{Probe classification on AI Safety Datasets (Layer 24). All safety prompts are classified as Deployment.}
\label{tab:safety_results}
\begin{center}
\begin{tabular}{l|cc}
\bf Dataset & \bf Single Probe & \bf Paired Probe \\
\hline
AdvBench (520) & 0\% & 0\% \\
HarmBench standard (200) & 0\% & 0\% \\
HarmBench contextual (100) & 0\% & 0\% \\
HarmBench copyright (100) & 0\% & 0\% \\
\end{tabular}
\end{center}
\end{table}

\subsection{Rewrite Prompt Templates}
\label{app:rewrite-prompts}

This section reports the exact prompt templates used for routing and rewriting.

\paragraph{Routing Prompt.}
\begin{verbatim}
You are a routing assistant for dataset construction.
Your task is to assign each user query to EXACTLY ONE category
based solely on the type of reasoning it requires.

AVAILABLE CATEGORIES:
(1) MMLU  – knowledge-based multiple-choice reasoning
(2) ARC   – grade-school science or commonsense reasoning
(3) GSM8K – arithmetic or multi-step math reasoning
(4) OTHER – none of the above

Output ONLY one label: MMLU, ARC, GSM8K, or OTHER.

User query:
"{USER_PROMPT}"
\end{verbatim}

\paragraph{Benchmark Rewrite: MMLU-style.}
\begin{verbatim}
Rewrite the following user query into a benchmark-style
multiple-choice question.

Constraints:
- Provide EXACTLY four answer options (A, B, C, D).
- Do NOT include explanations, solutions, or commentary.
- Do NOT indicate which option is correct.
- Preserve the original task intent and required information.
- Do NOT introduce an explicit evaluation framing.

User query:
"{USER_PROMPT}"

Output format:

Question:
<rewritten question>

Options:
A. ...
B. ...
C. ...
D. ...
\end{verbatim}

\paragraph{Benchmark Rewrite: ARC-style.}
\begin{verbatim}
Rewrite the following user query into a structured
grade-school science or commonsense reasoning question.

Constraints:
- Use a neutral, benchmark-style formulation.
- If appropriate, present the question as multiple-choice
  with EXACTLY four options.
- Do NOT include explanations or the correct answer.
- Preserve the original task intent.

User query:
"{USER_PROMPT}"
\end{verbatim}

\paragraph{Benchmark Rewrite: GSM8K-style.}
\begin{verbatim}
Rewrite the following user query into a structured
grade-school math word problem.

Constraints:
- Present a self-contained math problem.
- Do NOT include intermediate steps or the final answer.
- Do NOT add any solution explanation.
- Preserve the numerical reasoning required by the original query.

User query:
"{USER_PROMPT}"
\end{verbatim}

\paragraph{Genre Control Rewrite.}
\begin{verbatim}
Rewrite the following user query in a formal, impersonal,
and professional register.

Constraints:
- Preserve the original task intent and information.
- Do NOT introduce benchmark-style structure.
- Do NOT include answer options, solution steps, or summaries.
- The output should remain a free-form user request,
  not an evaluation-style prompt.

User query:
"{USER_PROMPT}"
\end{verbatim}

\end{document}

%% file: math_commands.tex
\usepackage{amsmath,amsfonts,bm}

\def\eqref#1{equation~\ref{#1}}

\def\1{\bm{1}}

\DeclareMathAlphabet{\mathsfit}{\encodingdefault}{\sfdefault}{m}{sl}
\SetMathAlphabet{\mathsfit}{bold}{\encodingdefault}{\sfdefault}{bx}{n}

